# Feature Space Renormalization for Semi-supervised Learning

Jun Sun, *IEEE Member*, Wancheng Zhang, Chao Zhou, Zhongjie Mao, Chao Li, Xiao-Jun Wu

***Abstract*—Semi-supervised learning (SSL) has been proven to be a powerful method for leveraging unlabeled data to alleviate models' dependence on large labeled datasets. The common framework among recent approaches is to train the model on a large amount of unlabeled data with consistency regularization to constrain the model predictions to be invariant to input perturbation. This paper proposes a feature space renormalization (FSR) mechanism for SSL, which imposes consistency on feature representations rather than on labels to enable the model to learn better discriminative features. In order to apply this mechanism to SSL, we design a dual-branch FSR module consisting of a dual-branch header and an FSR block. This module can be seamlessly plugged and played into existing SSL frameworks to enhance the performance of the base SSL. The experimental results show that our proposed FSR module helps the base SSL framework (e.g. CRMatch and FreeMatch), achieve better performance on a variety of standard SSL benchmark datasets, without incurring additional overhead in terms of computation time and GPU memory.**

## I. INTRODUCTION

DEEP neural networks (DNNs) have achieved great success in performing computer vision tasks. Such success is partially attributed to the existence of large labeled datasets, for example, ImageNet [1], COCO [2] and other datasets in various fields of computer vision. This means that training DNNs on large labeled datasets can yield better performance with supervised learning [3]-[5]. For many tasks, however, it is challenging to gather extensive accurately labeled data. One reason is that incorrect labels may be added due to subjective factors, and the other is that the cost of labeling can be extremely high since it must be done by professionals manually. For instance, medical image datasets must be labeled by experienced doctors [6]. In contrast, it is much easier to obtain unlabeled data in most tasks.

This work was supported in part by the National Natural Science Foundation of China under Grants 61672263 and 62272202. *(Corresponding author: Jun Sun).*

Jun Sun is with the School of Artificial Intelligence and Computer Science, Jiangnan University, Wuxi, Jiangsu 214122, China (e-mail: junsun@ jiangnan.edu.cn).

Wancheng Zhang, Chao Zhou, Zhongjie Mao, Chao Li are with the School of Artificial Intelligence and Computer Science, Jiangnan University, Wuxi, Jiangsu 214122, China (e-mail: wanchengzhang@stu.jiangnan.edu.cn).

Xiaojun Wu is with the School of Artificial Intelligence and Computer Science, Jiangnan University, Wuxi, Jiangsu 214122, China (e-mail: wu-xiaojun@jiangnan.edu.cn).

A powerful method of training models on a large amount of data without all the labels for the dataset is semi-supervised learning (SSL) [7], which can reduce the demand for labeled data by making the best use of unlabeled data. In many recent SSL approaches, a regularization term is added to the loss function based on unlabeled data to enable the model to generate better predictions for unseen data. There are mainly two choices for this loss term. The first is consistency regularization, which encourages a model to produce the same prediction when the input of the model is randomly modified [8]-[11], [26]. The second is pseudo-labeling, which employs confident category predictions for unlabeled data as pseudo labels to train the model [12], [13]. The current core framework of SSL generally integrates both of these two approaches to obtain better performance [14], [15].

In this paper, we propose a feature space renormalization (FSR) mechanism for SSL based on group representation theory, which implements consistency at the feature level to improve the model's ability to learn a good feature representation of unlabeled data. Specifically, FSR works as a lightweight dual-branch module before the classifier. The features of weakly augmented unlabeled samples are picked out from one branch and those of strongly augmented ones from the other branch, and they are together subject to FSR constraints during model training. The dual-branch FSR module can be seamlessly integrated into the existing SSL pipelines, e.g. CRMatch [16] and FreeMatch [17], yielding improved performance for the base SLL framework.

Fig.1 illustrates the main difference between the standard framework in recent SSL approaches (i.e. the base SSL framework) and the SSL framework with the dual-branch FSR module, named the FSR-augmented SSL framework. During model training in the base SSL approach, the input unlabeled image, along with the labeled one, is first preprocessed by different transformation techniques (e.g. weak augmentation and strong augmentation) to yield two or more images from different views. Then, the backbone model generates different predictions for these images in the label space. Finally, one of these predictions is regarded as the artificial label, and the other predictions are employed to predict this artificial label. These transformation techniques are either data augmentation methods [10],[14]-[20] or methods employing adversarial samples [22]. In this work, following the recent SSL approaches, we apply two different data augmentation methods to transform the input images. They are weak augmentation methods and strong augmentation methods, similar to those used in UDA [14].

As shown in Fig. 1, the proposed FSR-augmented SSL framework is different from the standard SSL framework in that the dual-branch FSR module is added between the backbone model and the classification layer. During model training, just as in the standard SSL framework, the unlabeled image is perturbed by weak augmentation and strong augmentation respectively, while the labeled image is transformed by weak augmentation. The resultant unlabeled and labeled images are input to the backbone model for feature extraction. Then, the feature samples output by the backbone model go through the dual-branch header with each branch having the same structure. These two branches reduce the dimensionality of feature samples and generate two different feature representation spaces with the same dimensionality. After that, the features of weakly augmented unlabeled samples and strongly augmented ones are picked up separately from either of the branches, and feature space renormalization is performed on these two feature representations. The loss term corresponding to feature space renormalization, on the basis of group representation theory, is derived from the condition for isomorphism between the covariance matrix groups corresponding to the feature representation spaces generated from weakly augmented images and strongly augmented ones. Meanwhile, pseudo-labeling and consistency regularization are also implemented for model training, just as in the standard SSL methods.

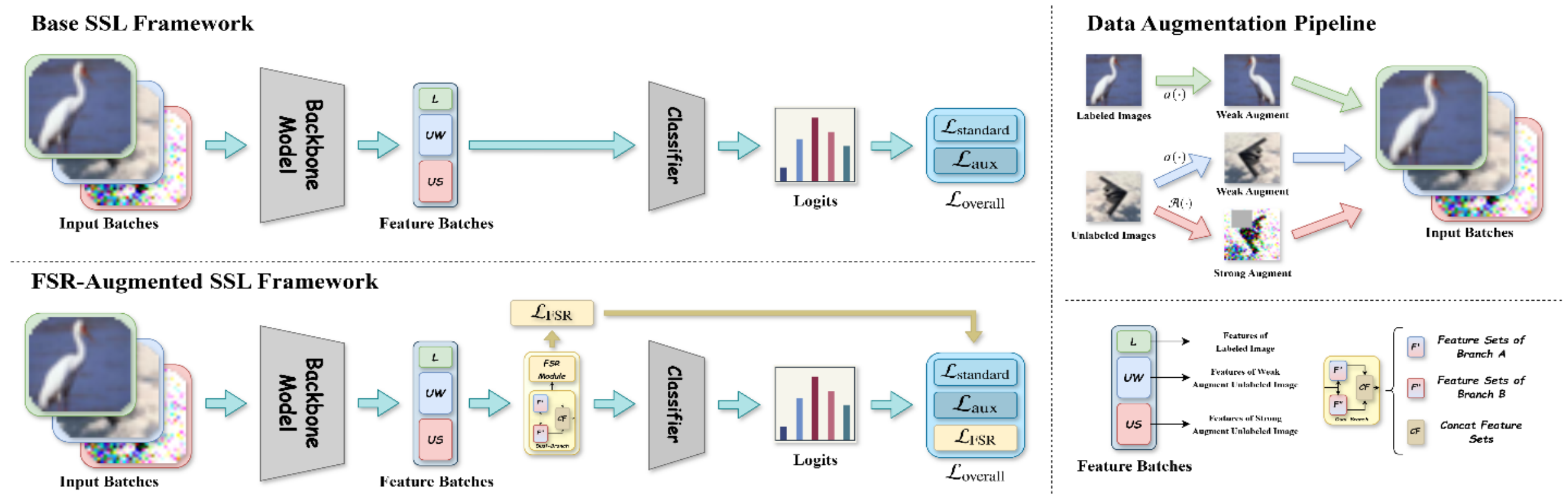

**Fig. 1.** The standard framework in recent SSL methods (named base SSL framework) and the proposed FSR-augmented SSL framework. The only difference between the two frameworks is that the latter includes a dual-branch FSR module between the backbone model and the classifier. Both frameworks use weak and strong augmentations, denoted as $a(\cdot)$ and $\mathcal{A}(\cdot)$, respectively.

The FSR-augmented SSL framework can obtain better performance on most commonly studied benchmarks for SSL due to our proposed feature space renormalization mechanism and the dual-branch FSR module, which are the main contributions of this paper. In particular, feature space renormalization, as the core of the FSR module, highlights the important role of feature generation and optimization in SSL and alleviates the learning burden on the subsequent classifier. It is an advancement in feature representation learning and is a substitute for or a complement to consistency regularization, which is widely used in existing SSL methods. Furthermore, the dual-branch FSR module is essentially a lightweight plug-and-play module that can be integrated into most of the existing SSL approaches, such as CRMatch and FreeMatch. As shown by our experimental results, with the help of the feature space renormalization mechanism, our proposed FSR-augmented SSL framework achieves generally better or comparable performance on the benchmark datasets compared to the corresponding base SSL framework and the other tested methods.

The paper is structured as follows. Section 2 reviews the related work of existing semi-supervised learning methods. In Section 3, the principles of the proposed FSR mechanism and the dual-branch FSR module for SSL are described. Section 4 presents the experimental results. Finally, the paper is concluded in the last section.

## II. Related Work

In this section, we review some existing methods used in SSL, which are currently state-of-the-art and mainly focus on improvements in consistency regularization and pseudo-labeling methods. Other methods, including transductive learning [23], [24], graph-based methods [25]-[27] and generative models [22], [28]-[30], are discussed briefly, and one can refer to the survey in [31] for a more comprehensive discussion of these methods.

For a $c$-class classification task, we define $\mathcal{X} = \{(\mathbf{X}_b, p_b) : b \in (1, \cdots, B)\}$ as the batches of labeled feature samples, where $\mathbf{X}_b$ is the $b$-th batch of labeled feature samples, $p_b$ are the labels corresponding to $\mathbf{X}_b$, $B$ is the number of sample batches, and $b$ denotes the sample batch number. In addition, we let $\mathcal{U} = \{\mathbf{U}_b : b \in (1, \cdots, B)\}$ be the batches of unlabeled feature samples and denote as $P_{model}(y|x; \theta)$ the prediction distribution for the categories by the model $p$ for input $x$ with parameters $\theta$.

### A. Consistency Regularization

Consistency has been employed in various loss functions for DNN training. These loss functions, including MSE [32], KL divergence [33], [34] and cross-entropy [35], are used to minimize the gap between the prediction and the label, which means that the prediction and the label should be as consistent as possible. Therefore, consistency is an intrinsic and essential objective in enabling DNNs to learn more effective features. In SSL, consistency regularization is carried out for unlabeled data, relying on the assumption that the model should output the same predictions when fed perturbed views of the same sample. A common perturbation technique is data augmentation [36], [37], which carries out image transformations while keeping the category semantics unaffected. In this work, we also employ data augmentation to add perturbations to unlabeled data. Formally, consistency regularization requires that an unlabeled sample $u_i$ should be classified into the same category as its corresponding augmented sample, i.e., $Aug(u_i)$.

According to the above description, most SSL approaches [14],[15] apply the loss term in equation (1) for a batch of unlabeled data:

$$\frac{1}{n}\sum_{i=1}^{n}||P_{model}(y|Aug_1(u_i);\theta) - P_{model}(y|Aug_2(u_i);\theta)||_2^2 \quad (1)$$

where $Aug_1(\cdot)$ and $Aug_2(\cdot)$ represent two different data augmentation operations, $u_i$ is the $i$-th sample in the batch, and $n$ is the batch size. For example, in the Mean Teacher method [9], $Aug_2(\cdot)$ in equation (1) is defined by the output of the model updated with a moving average approach, and this operation provides a relatively stable feature representation so that it can significantly improve the classification performance. The virtual adversarial training (VAT) method [22] determines an additional perturbation direction, which is applied to the input data, thereby maximizing the change in the distribution of the predicted category. It is essentially a training method with $Aug_2(\cdot)$ in equation (1) being an adversarial transformation. In MixMatch [10], the MixUp method is used for data augmentation. Based on MixMatch, ReMixMatch was proposed in [11] with two new data augmentation techniques, namely, distribution alignment and augmentation anchoring. In [14], UDA used a weakly augmented sample to generate an artificial label and then enforced the prediction so that the strongly augmented sample was consistent with the artificial label. In UDA, the artificial label is also sharpened to encourage the model to produce predictions with high confidence. FixMatch [15], a simplified version of UDA [14], combines pseudo-labeling and consistency regularization but without the sharpening and training signal annealing in UDA [14].

Simply imposing consistency constraints at the decision layer causes the information in the feature representation layer to be ignored. Thus, researchers have turned to investigate how to implement consistency regularization at the feature representation layer before the classifier to learn more discriminative and structured features. For example, in SimMatch [20], consistency regularization is carried out at both the semantic level (i.e. the decision level) and instance level (i.e. the feature level). SimMatch-V2 [22] formulates various consistency regularizations between labeled and unlabeled data from the graph perspective, with the augmented view of a sample regarded as a node. In CRMatch [16], the feature distance between strongly and weakly augmented samples is selectively increased in the feature space to help the model learn more robust discriminative boundaries.

In contrast to the existing consistency regularization methods, our proposed feature space renormalization mechanism renormalizes the feature representation of strongly augmented unlabeled samples with that of weakly augmented ones to make the two feature spaces as homeomorphic as possible.

*B. Entropy Minimization and Pseudo-labeling*

A common assumption in SSL is that the decision boundary of a classifier should not pass through the high-density regions of the marginal data distribution [10]. Entropy minimization [38], [39] can satisfy this assumption by encouraging the model to output low-entropy predictions on unlabeled data. This method can be applied explicitly with a loss term that minimizes the entropy of $P_{model}(y|x;\theta)$ for unlabeled data. Another approach that can also satisfy this assumption is pseudo-labeling, which employs the model itself to obtain pseudo labels for unlabeled data [40], [41]. Moreover, pseudo-labeling can perform entropy minimization implicitly by using "hard" labels when the largest category probability exceeds a predefined threshold [12]. By defining $q_i = P_{model}(y|u_i)$, we can obtain the loss function for unlabeled data as

$$\frac{1}{n}\sum_{i=1}^{n}\mathbb{1}(\max(q_i) \geq \eta)H(\hat{q}_i, q_i) \quad (2)$$

where $\hat{q}_i = \arg\max(q_i)$, $\eta$ is a hyperparameter called the threshold, $H(\cdot)$ is the standard cross-entropy loss function, and $\mathbb{1}(\cdot)$ is the mask function. In addition, the $\arg\max(\cdot)$ operation is applied to the output of the model to yield a valid "one-hot" probability distribution. It was shown that the loss in this form, combined with VAT, can lead to competitive classification performance [26].

Some self-training methods also involve entropy minimization. For instance, in MixMatch [10], a sharpening function is utilized on the target distribution for unlabeled data by implicitly performing entropy minimization. FlexMatch [42] employs curriculum pseudo-labeling (CPL), a curriculum learning approach exploiting unlabeled data according to the learning status of the model. Furthermore, the core of CPL is that it flexibly adjusts the thresholds for different classes at each training iteration, allowing informative unlabeled data and their pseudo-labels to pass the threshold. In FeatMatch [43], a novel learned feature-based refinement and augmentation method is proposed to produce a varied set of good transformations.

Many recent works on SSL have mainly focused on optimizing strategies for pseudo-label selection. Various flexible threshold adjustment mechanisms have been developed to replace fixed thresholds for pseudo-labeling. For example, Meta Pseudo Labels (MPL) couples a teacher–student architecture with a meta-learning strategy to adapt the student's pseudo-labels [44]. In Dash [45], the threshold is dynamically lowered during model training. This strategy is based on the assumption that the model becomes more reliable in the later stages of training. AdaMatch [46] designs a relative confidence threshold, which is adaptively adjusted based on the average prediction of the model on weakly augmented samples. SoftMatch [47] further abandons hard thresholding and uses a Gaussian function to assign smooth weights to pseudo-labels with different confidence levels. FlexMatch [42] and FreeMatch [17] further develop more refined adaptive threshold strategies. FreeMatch, in particular, adaptively adjusts the confidence threshold for each category based on the model's learning state, significantly improving performance in scenarios with very few labeled samples. Furthermore, to utilize those unselected low-confidence samples, FullMatch [48] employs adaptive negative learning, which allows all the unlabeled samples, regardless of their confidence levels, to provide meaningful supervisory information for training.

### *C. Other Methods Used in Semi-supervised Learning*

In addition to the above two kinds of methods commonly used in SSL, researchers have explored various methods in different fields to improve the performance of SSL, including deep clustering and self-supervised learning. For example, Lerner et al. [49] proposed combining a clustering algorithm with deep SSL methods so that the SSL methods can benefit from real unsupervised learning and are not affected by a small number of possibly uncharacteristic training points. In [50], Wallin et al. designed a novel consistency regularization approach based on self-supervised learning and combined it with the pseudo-labeling technique so that the model could make full use of all unlabeled data during the training process. In CoMatch [51], the learned low-dimensional embeddings impose a smoothness constraint on the class probabilities to improve the pseudo-labels, whereas the pseudo-labels regularize the structure of the embeddings through graph-based contrastive learning.

## III. THE FSR-AUGMENTED SSL FRAMEWORK

In this section, we describe in detail the principle and implementation of our proposed feature space renormalization mechanism and the FSR module for SSL. First, we present the feature representation based on group representation theory, which is the theoretical foundation of the proposed feature space renormalization mechanism. Second, the principle of feature space renormalization is described. Then, we describe in detail the SSL framework with the dual-branch FSR module. Next, we illustrate how to implement feature space renormalization for the FSR-augmented SSL framework. After that, the pseudo-labeling method used in SSL model training is presented. Finally, we provide the total loss function for model training in the FSR-augmented SSL framework.

In the rest of this section, we still use the notation and definitions in Section 2, and we define two types of augmentations, namely, strong augmentation $\mathcal{A}(\cdot)$ and weak augmentation $\alpha(\cdot)$. For clarity, we show the structure of FSR-augmented SSL, particularly the dual-branch FSR module, in Fig. 2, which is a refinement of the structure in Fig. 1.

### *A. Feature Representation Based on Group Representation Theory*

In general, a data sample is represented as a feature vector in a $D$-dimensional Euclidean space $\mathcal{H}$, which is called the feature representation space or data space. Mathematically, $\mathcal{H}$ can be regarded as a separable topological space (i.e., Hausdorff space). Therefore, we can represent a set of feature samples as a real $N \times D$ matrix, where $N$ is the sample size and $D$ is the number of features, namely, the dimension of the feature representation space $\mathcal{H}$. From a statistical perspective, the general characteristic of the data space is the distribution characteristic of the samples in the space. This characteristic should reflect the symmetry of the space, including translation invariance and rotation invariance. Since the covariance matrix of a given set of samples has a certain invariant characteristic under translation and rotation, it can reflect the symmetry of the data space and therefore represents the fundamental distribution characteristic of the data space.

A covariance matrix is a real $D \times D$ symmetric matrix, and the covariance matrices of all the different feature sample sets in the same feature representation space $\mathcal{H}$ are assembled to compose a subgroup $\mathbf{\Sigma}_D$ of the matrix group $GL_D$, where $GL_D$ is a general linear group composed of all the $D \times D$ real invertible matrices [52]. We call $\mathbf{\Sigma}_D$ the covariance matrix group of $\mathcal{H}$. Furthermore, $\mathbf{\Sigma}_D$ is also a topological group since the multiplication and inverse operations on it are essentially continuous maps [52]. Therefore, we can establish the corresponding relation between the Hausdorff space $\mathcal{H}$ and its topological group $\mathbf{\Sigma}_D$; that is, $\mathbf{\Sigma}_D$ can represent $\mathcal{H}$.

For two $D$-dimensional feature representation spaces $\mathcal{H}$ and $\mathcal{H}'$, they are said to be topologically the same if they are homeomorphic [53]. Equivalently, a homeomorphism between $\mathcal{H}$ and $\mathcal{H}'$ means that their corresponding covariance matrix groups, $\mathbf{\Sigma}_D$ and $\mathbf{\Sigma}_D{}'$, are isomorphic. According to group representation theory, the isomorphism between $\mathbf{\Sigma}_D$ and $\mathbf{\Sigma}_D{}'$ requires that for any given sample set $S$, whose feature sample set represented in $\mathcal{H}$ and $\mathcal{H}'$ are $\mathbf{X}_S$ and $\mathbf{X}'_S$, respectively, there exists an invertible matrix $\mathbf{P} \in GL_D$ such that

$$\boldsymbol{\sigma}_S = \mathbf{P}^{-1}\boldsymbol{\sigma}'_S\mathbf{P}, \tag{3}$$

where $\boldsymbol{\sigma}_S \in \mathbf{\Sigma}_D$ and $\boldsymbol{\sigma}'_S \in \mathbf{\Sigma}_D{}'$ are the covariance matrices of $\mathbf{X}_S$ and $\mathbf{X}'_S$, respectively. Equation (3) essentially means that $\boldsymbol{\sigma}_S$ and $\boldsymbol{\sigma}'_S$ are equivalent matrices. From this equation, we can derive

$$\mathrm{tr}(\boldsymbol{\sigma}_S) = \mathrm{tr}(\mathbf{P}^{-1}\boldsymbol{\sigma}'_S\mathbf{P}) = \mathrm{tr}(\boldsymbol{\sigma}'_S), \tag{4}$$

where $\mathrm{tr}(\cdot)$ denotes the trace of the matrix. In group representation theory [54], $\mathrm{tr}(\cdot)$ is known as the character function for a matrix group, which can characterize the feature representation space to a certain extent.

From the above description, if two feature representation spaces $\mathcal{H}$ and $\mathcal{H}'$ are homeomorphic, or equivalently their corresponding covariance matrix groups $\mathbf{\Sigma}_D$ and $\mathbf{\Sigma}_D{}'$ are isomorphic, then their character functions are the same. However, the inverse proposition does not hold; that is, we cannot derive equation (3) from equation (4) since $\mathrm{tr}(\cdot)$ is a group homomorphism but not necessarily a bijection [54].

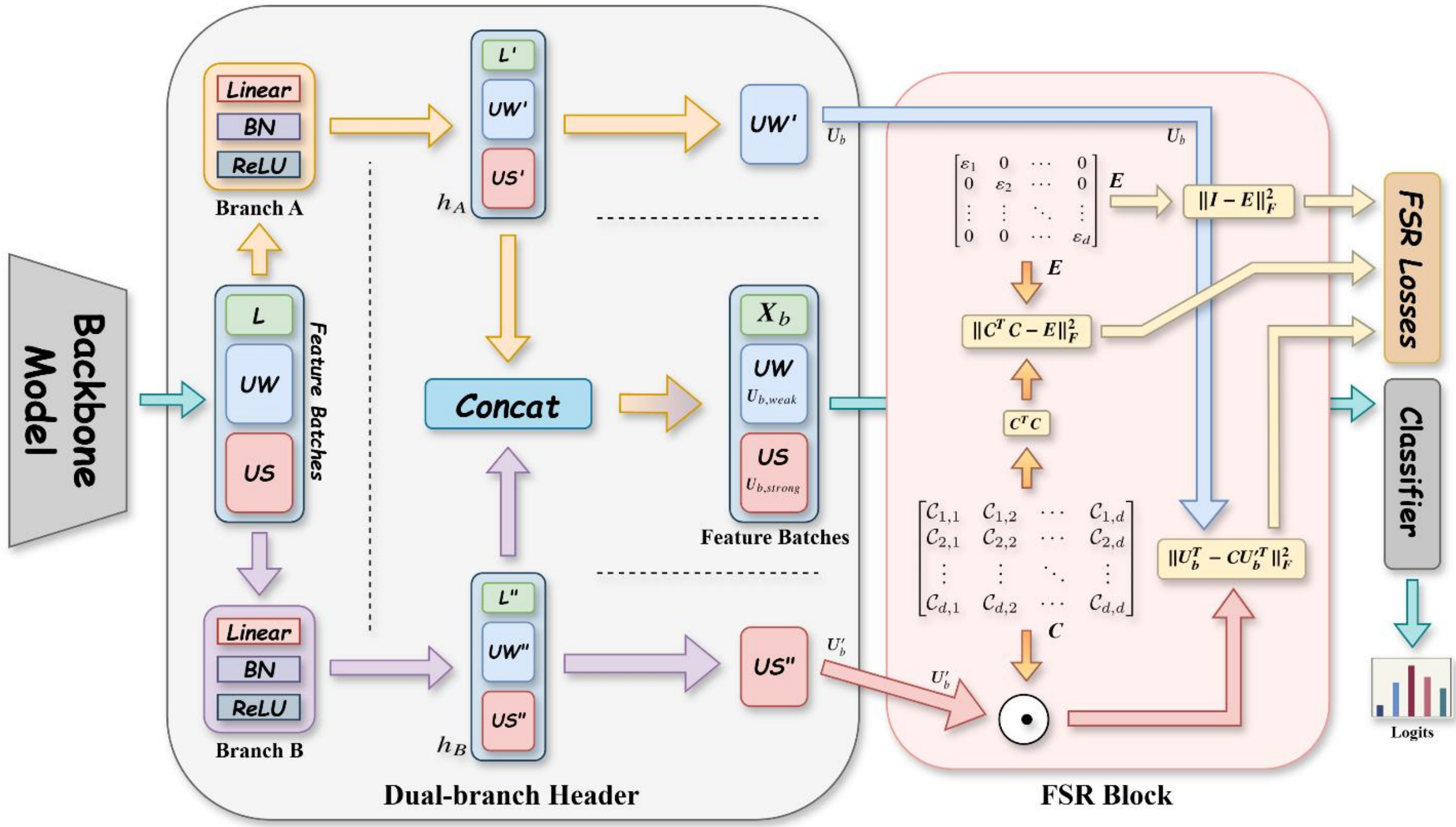

**Fig. 2.** The proposed dual-branch module in the FSR-augmented SSL framework. It is composed of two parts, a dual branch header (DBH) and an FSR block. The DBH has two parallel and structurally identical branches, each of which has a linear layer, followed by a batch normalization (BN) layer and then by a LeakyRelU activation. The features output by both branches are concatenated and then input to the classifier in inference during model training or testing. The features of weakly augmented unlabeled image output by branch *A* and those of corresponding strongly augmented one output by branch *B* are picked out separately to perform FSR during model training.

### *B. Feature Space Renormalization*

According to the above analysis, we introduce our proposed feature space renormalization mechanism as follows.

Feature space renormalization is carried out on the features to make the two feature representation spaces $\mathcal{H}$ and $\mathcal{H}'$ homeomorphic, or equivalently to make their corresponding covariance matrix groups $\mathbf{\Sigma}_D$ and $\mathbf{\Sigma}_D{}'$ isomorphic. Without loss of generality, we denote $\mathbf{X}_S$ and $\mathbf{X}'_S$ as the centralized feature sample sets for the data sample set $S$ in the feature representation spaces $\mathcal{H}$ and $\mathcal{H}'$, respectively. Here, centralization means that each dimension of each sample vector is decreased by the average value of the corresponding dimension. Therefore, the covariance matrices for $\mathbf{X}_S$ and $\mathbf{X}'_S$ are given by

$$\boldsymbol{\sigma}_S = \mathbf{X}_S^T \mathbf{X}_S \tag{5}$$

and

$$\boldsymbol{\sigma}'_S = \mathbf{X}_S'^T \mathbf{X}'_S. \tag{6}$$

To make $\mathbf{\Sigma}_D$ and $\mathbf{\Sigma}_D{}'$ isomorphic or $\boldsymbol{\sigma}_S$ and $\boldsymbol{\sigma}'_S$ equivalent, we could use equation (4) as the constraint condition, that is, minimize $|\mathrm{tr}(\boldsymbol{\sigma}_S) - \mathrm{tr}(\boldsymbol{\sigma}'_S)|$ during feature learning. However, as addressed above, equation (4) cannot guarantee the equivalence between $\boldsymbol{\sigma}_S$ and $\boldsymbol{\sigma}'_S$, and moreover, it can lead to information loss during feature learning if only equivalence between the character functions is required. Hence, we have to employ equation (3) to make $\boldsymbol{\sigma}_S$ and $\boldsymbol{\sigma}'_S$ equivalent. To this end, we assume there exists a mapping $\boldsymbol{C}: \mathcal{H}' \rightarrow \mathcal{H}$ such that

$$\mathbf{X}_S^T = \boldsymbol{C}\mathbf{X}_S'^T \tag{7}$$

where $\boldsymbol{C}$ is a $D \times D$ real matrix. Consequently, from equations (5) and (7), we can obtain

$$\boldsymbol{\sigma}_S = \mathbf{X}_S^T \mathbf{X}_S = \boldsymbol{C}\mathbf{X}_S'^T \mathbf{X}'_S \boldsymbol{C}^T. \tag{8}$$

By substituting (6) into (8), we have

$$\boldsymbol{\sigma}_S = \boldsymbol{C}\boldsymbol{\sigma}'_S\boldsymbol{C}^T. \tag{9}$$

Comparing equation (9) with equation (3), we find that if $\boldsymbol{C}^T = \mathbf{P}$ and $\boldsymbol{C} = \mathbf{P}^{-1}$, or equivalently, if

$$\boldsymbol{C}^T\boldsymbol{C} = \boldsymbol{I}_D, \tag{10}$$

then $\boldsymbol{\sigma}'_S$ is equivalent to $\boldsymbol{\sigma}_S$ so that the two feature representation spaces are homeomorphic or the same. Here, $\boldsymbol{I}_D$ is the $D \times D$ identity matrix, and equation (10) indicates that $\boldsymbol{C}$ should be an orthogonal matrix to satisfy equation (3). Therefore, equations (7) and (10) together can be employed as the constraints during the training process, and all the elements of $\boldsymbol{C}$ are learnable parameters. Such a mechanism using equations (7) and (10) as the constraint to ensure the consistency at the feature level is called feature space renormalization (FSR). In the rest of this section, we will further describe how to implement feature space renormalization for our proposed FSR-augmented SSL.

*C. SSL with the Dual-branch FSR Module*

In the SSL approaches, a DNN model is employed as the backbone model to generate the feature representation for the given samples, both labeled samples and unlabeled ones after weak augmentation or strong augmentation. During the model training for the standard SSL framework, the resultant features are input to classification layer for implementing pseudo-labeling and consistency regularization for unlabeled samples or obtaining cross entropy for labeled samples. Instead, during the model training process in the SSL integrated with the proposed FSR module, i.e. the FSR-augmented SSL, the output features by the backbone model go through a dual-branch FSR module, which consist of a dual-branch header and an FSR block. The role of the dual-branch header is to perform feature dimensionality reduction and generate two different feature representation spaces with the same dimensionality for the feature samples, which are subsequently subjected to feature space renormalization. The FSR block only performs feature space renormalization on features of strongly augmented images and weakly augmented ones represented in the two different feature representation spaces $\mathcal{H}$ and $\mathcal{H}'$ in the header during model training, not participating in inference for model training or testing.

Specifically, as shown in Fig.2, the dual-branch header appends two parallel and structurally identical branches to the backbone model. Each branch has a linear layer, followed by a batch normalization (BN) layer with the momentum being 0.001 and then by a LeakyRelU activation with the negative being 0.1. The hyperparamters in the header are set to be the same as those in the backbone network to ensure that normalization statistics and activation dynamics remain consistent throughout the network. Formally, let $\boldsymbol{z} \in \mathbb{R}^D$ reprensents the feature vector generated by the backbone network for an input image, and the outputs by the two branches are given respectively by

$$\boldsymbol{h}_A = \psi(BN(\boldsymbol{W}_A\mathbf{z})) \tag{11}$$

and

$$\boldsymbol{h}_B = \psi(BN(\boldsymbol{W}_B\mathbf{z})) \tag{12}$$

where $\boldsymbol{W}_A, \boldsymbol{W}_B \in \mathbb{R}^{d\times D}$ are the learnable weight matrices of linear layers in branch $A$ and branch $B$, respectively. $BN(\cdot)$ is the batch normalization operator, and $\psi(\cdot)$ denotes the LeakyReLU activation function. Here, $d = D/2$, which means that each branch reduces the feature vector dimension by half. The purpose of dimensionality reduction by each branch is to lighten the burden of learning for the FSR module during model training.

In inference during model training or testing, the feature vectors output by the two branches, i.e. $\boldsymbol{h}_A$ and $\boldsymbol{h}_B$, are concatenated to form the fused feature vector:

$$\boldsymbol{x} = [\boldsymbol{h}_A||\boldsymbol{h}_B] \in \mathbb{R}^D \tag{13}$$

which is then input to the classifier for prediction, as shown in Fig 2. Since $d = D/2$, the dimension of $\boldsymbol{x}$ is $D$. This means that the feature vector input to the classifier has the same dimensionality as the one to the classifier in the standard SSL architecture without the dual-branch FSR module. Therefore, the computational overhead incurred by the FSR module can be negligible.

During model training, each mini-batch of training samples consist of weakly augmented labeled images, weakly augmented unlabeled images and strongly augmented unlabeled images. The training samples are fed to the backbone model and the dual-branch header. Each branch of the header output the feature vectors of these images. The feature vector of the weakly augmented unlabeled image output by one branch and that of the corresponding strongly augmented one by the other branch are picked out separately to perform feature space renormalization. Specifically, for the $i$-th unlabeled training sample in the $b$-th batch, we denote as $\boldsymbol{h}_{A,weak}^{i,b}$ the feature vector yielded by branch A for the weakly augmented image, and $\boldsymbol{h}_{B,strong}^{i,b}$ the feature vector yield by branch $B$ for the corresponding strongly augmented image. Hence, the $b$-th batch of weakly augmented feature samples produced by branch $A$ and the corresponding strongly augmented feature samples by branch $B$ can be represented respectively by

$$\mathbf{U}_b = \left[\boldsymbol{h}_{A,weak}^{1,b}, \boldsymbol{h}_{A,weak}^{2,b}, \cdots, \boldsymbol{h}_{A,weak}^{n,b}\right] \tag{14}$$

and

$$\mathbf{U}'_b = \left[\boldsymbol{h}_{B,strong}^{1,b}, \boldsymbol{h}_{B,strong}^{2,b}, \cdots, \boldsymbol{h}_{B,strong}^{n,b}\right] \tag{15}$$

on which feature space renormalization is performed, as shown by the FSR block in Fig. 2. Here, $n$ is the number of the unlabeled sample pairs in the batch. The implementation details of FSR will be described in the next subsection. It should be noted that the FSR block does not work in inference.

*D. Implementation of FSR*

In this subsection, we will illustrate in detail how to implement feature space renormalization in our FSR-augmented SSL and then deduce its loss function.

In the FSR-augmented SSL architecture, feature space renormalization is carried out on each batch of unlabeled samples during the training process. For the $b$-th batch of unlabeled samples, strong augmentation and weak augmentation are carried out on the unlabeled sample batch, and the resultant sample batches are fed to the backbone model, and then go through the two branches in the header of the FSR module, respectively. Hereafter, the weakly augmented feature sample batch produced by branch $A$ and the strongly augmented feature sample batch by branch $B$, namely, $\mathbf{U}_b \in \mathbb{R}^{n\times d}$ and $\mathbf{U}'_b \in \mathbb{R}^{n\times d}$, are subjected to feature space renormalization. Here, $\mathbf{U}_b$ and $\mathbf{U}'_b$ are centralized, and they are represented respectively in two different feature representation spaces, $\mathcal{H}$ and $\mathcal{H}'$, both with dimension $d$.

Following the feature space renormalization mechanism proposed in subsection B, we assume that there exists a mapping $\boldsymbol{C}: \mathcal{H}' \to \mathcal{H}$ such that

$$\mathbf{U}_b^T = \boldsymbol{C}\mathbf{U}_b'^T, \tag{16}$$

as in equation (7). Thus, the condition for an isomorphism existing between the covariance matrices of $\mathcal{H}$ and $\mathcal{H}'$ is

$$\boldsymbol{C}^T\boldsymbol{C} = \boldsymbol{I}_d, \tag{17}$$

which is identical to equation (10) except that the dimension of the feature representation spaces is $d$.

However, it should be noted here that there is much difference between the weakly augmented sample batch and the corresponding strongly augmented sample batch. Therefore, although the network from backbone to branch $A$ and the one from backbone to branch $B$ have the same structure, it is hardly reasonable to require that $\mathcal{H}$ and $\mathcal{H}'$ are completely homeomorphic or exactly the same. In this sense, the condition for feature space renormalization in equation (17) is too strict. In fact, during model training, it is only need to require $\mathcal{H}$ and $\mathcal{H}'$ to be as homeomorphic as possible, or say, as similar as possible.

As such, it is reasonable to use a weaker condition to replace the one in equation (17). A feasible scheme of weakening the condition in equation (17) is to not require the orthogonality of $\boldsymbol{C}^T$ or $\boldsymbol{C}$. As a result, the right side of equation (17) should not be the identity matrix. Considering that $\boldsymbol{C}^T\boldsymbol{C}$ is a real symmetric matrix and the number of parameters to be learned during the training process should be as small as possible, we change equation (17) to

$$\begin{aligned}\boldsymbol{C}^T\boldsymbol{C} &= \boldsymbol{I}_d\,\mathrm{diag}(\varepsilon_1\varepsilon_2, \cdots, \varepsilon_d) \\ &= \mathrm{diag}(\varepsilon_1, \varepsilon_2, \cdots, \varepsilon_d) = \begin{pmatrix} \varepsilon_1 & & \\ & \ddots & \\ & & \varepsilon_d \end{pmatrix} = \boldsymbol{E},\end{aligned} \tag{18}$$

where $\boldsymbol{E} = \mathrm{diag}(\varepsilon_1, \varepsilon_2, \cdots, \varepsilon_d)$ is a diagonal matrix and $0 \le \varepsilon_j \le 1$ ($1 \le j \le d$) is called the tolerance in the $j$-th dimension.

Hence, we obtain the feature space renormalization conditions for SSL as equations (16) and (18) and rewrite them together as

$$\begin{gathered}\mathbf{U}_b^T = \boldsymbol{C}\mathbf{U}_b'^T, \\ \boldsymbol{C}^T\boldsymbol{C} = \mathrm{diag}(\varepsilon_1, \varepsilon_2, \cdots, \varepsilon_d),\end{gathered} \tag{19}$$

which are thus employed as the constraints for feature space renormalization during the training process. Here, $\boldsymbol{C}$ and $\varepsilon_j$ ($1 \le j \le d$) are the parameters to be learned during the model training.

Furthermore, we should also exert regularization on all the $\varepsilon_j$ ($1 \le j \le d$) to make $\boldsymbol{E}$ as close as $\boldsymbol{I}_d$. Thus, the regularization term can be

$$\|\boldsymbol{I}_d - \boldsymbol{E}\|_F^2 = \sum_{j=1}^{d}(1-\varepsilon_j)^2 \tag{20}$$

where $\|\cdot\|_F^2$ is the square of the Frobenius norm. Essentially, minimization of $\|\boldsymbol{I}_d - \boldsymbol{E}\|_F^2$, which also minimizes $|\mathrm{tr}(\mathbf{U}_b^T\boldsymbol{U}_b) - \mathrm{tr}(\mathbf{U}_b'^T\boldsymbol{U}'_b)|$, together with the constraints in equation (19), can achieve the goal of making $\mathcal{H}$ and $\mathcal{H}'$ as homeomorphic as possible.

With the dual-branch FSR module integrated in the SSL, the training of models is subject to the feature space renormalization constraints in equations (19) and the regularization term for FSR in equation (20). The equality constraints in (19) can be converted to penalty functions, and thus the loss term for FSR is given by

$$\begin{gathered}\boldsymbol{\mathcal{L}}_{\mathrm{FSR}} = \|\mathbf{U}_b^T - \boldsymbol{C}\mathbf{U}_b'^T\|_F^2 \\ +\lambda_b\|\boldsymbol{C}^T\boldsymbol{C} - \boldsymbol{E}\|_F^2 + \lambda_r\|\boldsymbol{I}_d - \boldsymbol{E}\|_F^2,\end{gathered} \tag{21}$$

where $\lambda_b$ and $\lambda_r$ are scalar hyperparameters, and they are set to be $\lambda_b = 0.01$ and $\lambda_r = 0.001$ throughout all the experiments in this work. Such parameter selection is justified by the sensitivity analysis in section IV.

*E. Pseudo-labeling*

For every mini batch of training samples, there are labeled samples and unlabeled samples. By weak augmentation and strong augmentation, $n_{lb}$ weakly augmented labeled samples and $n$ pairs of weakly augmented and strong augmented unlabeled samples are generated after balanced sampling, and then input to the model. After the training sample batch goes through the backbone and the dual-branch header, three concatenated feature sample sets are yielded, including labeled feature sample set $\mathbf{X}_b \in \mathbb{R}^{n_{lb} \times D}$ with the labels $\{y_i\}_{i=1}^{n_{lb}}$, the weakly augmented unlabeled feature sample set $\mathbf{U}_{b,weak} \in \mathbb{R}^{n \times D}$, and the strongly augmented unlabeled feature sample set $\mathbf{U}_{b,strong} \in \mathbb{R}^{n \times D}$. It should be noted that the latter two feature sets, obtained by concatenation in equation (13), are different from $\mathbf{U}_b$ and $\mathbf{U}_b'$, which are used to implement feature space renormalization. The classifier maps each feature vector to a prediction distribution.

We obtain the pseudo labels for the weakly augmented unlabeled training samples from the model, and use these labels to compute the standard cross-entropy with the predictions obtained from the model for the corresponding strongly augmented unlabeled training samples. Here, the model means “backbone model + dual-branch header + classifier”. Specifically, we obtain the prediction distribution of categories $q_i = P_{\text{model}}(y|a(u_i))$ by applying the model to the weakly augmented unlabeled samples and then obtain $\hat{q}_i = \arg\max(q_i)$. To learn better discrimination, we employ $\hat{q}_i$ as the artificial label when it is higher than a specified confidence threshold. Finally, we can obtain the loss function for pseudo-labeling $\mathcal{L}_{pl}$ as

$$\mathcal{L}_{\text{pl}} = \frac{1}{n}\sum_{i=1}^{n} \mathbb{1}(\max(q_i) > \eta)\, H\left(\hat{q}_i, P_{model}\big(y\big|\mathcal{A}(u_i)\big)\right), \quad (22)$$

where $\eta$ is a scalar hyper-parameter denoting the confidence threshold.

*F. The Loss Function*

For weakly augmented labeled training samples, the supervised training loss is the standard cross-entropy loss given by

$$\mathcal{L}_{\text{sup}} = \frac{1}{n_{lb}}\sum_{i=1}^{n_{lb}} H(p_i, P_{model}(a(x_i))). \quad (23)$$

The loss terms in equations (22) and (23) together constitutes the loss function for the standard SSL:

$$\mathcal{L}_{\text{standard}} = \lambda_u \mathcal{L}_{\text{pl}} + \mathcal{L}_{\text{sup}} \quad (24)$$

where $\lambda_u$ is a scalar hyperparameter.

Since our proposed dual-branch FSR module is inserted in the existing SSL framework as a plug-and-play module, the overall per-mini-batch loss function becomes

$$\mathcal{L}_{\text{overall}} = \underbrace{\mathcal{L}_{\text{standard}} + \mathcal{L}_{\text{aux}}}_{\text{Base SSL}} + \underbrace{\mathcal{L}_{\text{FSR}}}_{\text{FSR Module}} \quad (25)$$

where $\mathcal{L}_{\text{aux}}$ is the additional loss term besides $\mathcal{L}_{\text{standard}}$ in the base SSL framework into which the dual-branch FSR module is plugged. It is thus method-specific. For example, if the base SSL is FreeMatch, $\mathcal{L}_{\text{aux}}$ is the term for entropy minimization; if CRMatch, it is for distribution alignment.

## IV. EXPERIMENTS

*A. Experimental Settings*

We systematically evaluated the performance of the proposed FSR module for SSL on several standard SSL datasets, including CIFAR-10 [55], CIFAR-100 [55], SVHN [56], and STL-10 [57]. Different numbers of labeled samples were used in the experiments on these datasets for the purpose of comprehensive performance evaluation. For CIFAR-10, 4, 25, and 400 labels per class were used respectively; for CIFAR-100 and SVHN, 4, 25, and 100 labels per class were utilized respectively. For STL-10, since too few labels can cause noticeable instability during model training, we adopted the conventional setting of 100 labels per class to ensure reliable training and fair comparison.

Since the proposed dual-branch FSR module works as a plug-and-play component in the SSL architecture, we selected two representative state-of-the-art SSL approaches, i.e. CRMatch [16] and FreeMatch [17], as the base SSL frameworks, and inserted the FSR module into them respectively, yielding two FSR-augmented SSL variants, i.e. CRMatch-FSR and FreeMatch-FSR. Performance comparisons were made between each base SSL approach and its FSR-augmented variant under identical training protocols. It should be emphasized that the FSR module only involves a lightweight dual-branch header and the FSR loss term (only for model training) with the remainder of each underlying SSL model kept untouched. The baseline models we chose for performance comparison, besides CRMatch [16], and FreeMatch [17], included Π-Model [58], Mean Teacher [9], FixMatch [15], Dash [45], MPL [44], SoftMatch [47], ReFixMatch [59], DoubleMatch [42], SimMatchV2 [21], NP-Match [60], FullMatch [48], and OTAMatch [61].

To ensure fair comparisons, we implemented all the approaches in the same codebase and employed the Wide ResNet network [62] as the backbone network for the purpose of fair comparison as recommended in [63]. In detail, Wide ResNet-28-2 was employed for the CIFAR-10 and SVHN datasets, Wide ResNet-28-8 for CIFAR-100, and Wide ResNet-37-2 for STL-10. The entire training protocols, including the optimizer, learning rate schedule and data preprocessing, were the same for the same dataset.

In the experiments on the benchmark datasets, we trained the model by using the SGD with a momentum of 0.9. The initial learning rate was set to 0.01, and the cosine scheduler, $\alpha = \alpha_0 \cos\left(\frac{7\pi k}{16K}\right)$,was used for the gradual decay of the learning rate. Here, $\alpha_0$ is the initial learning rate, $k$ is the current iteration number, $K = 2^{20}$ is the total number of iterations.

For data augmentation, the weak augmentation method that solely consists of flip-and-shift transformations was employed, that is, the images were flipped horizontally with a probability of 0.5 followed by a random translation with a maximum distance of 0.125 of the image height; the strong augmentation used was RandAugment [64] followed by Cutout [18].

All the experiments were conducted on two Intel® Xeon Platinum 8336C CPUs running Ubuntu 22.04.5 LTS, equipped

with a single NVIDIA RTX 4090 GPU. The software stack consists of Python 3.12.7 and PyTorch 2.5.1 with the torch.compile feature and automatic mixed precision (AMP) enabled. No multi-GPU data parallelism was used.

The experimental results involve those of performance comparison, the comparison of training costs, the ablation study, and sensitivity analysis for hyperparameters. Each method was trained multiple times on every dataset with different label budgets, and the mean and standard deviation of the top-1 error rates were reported.

TABLE I

MEAN ERROR RATES (%) AND STANDARD DEVIATIONS ON CIFAR-10 AND CIFAR-100 DATASETS WITH DIFFERENT LABEL BUDGETS. THE BEST RESULTS ARE HIGHLIGHTED IN BOLD AND THE SECOND-BEST RESULTS ARE UNDERLINED

| Dataset | CIFAR-10 | | | CIFAR-100 | | |
|---|---|---|---|---|---|---|
| | 40 labels | 250 labels | 4000 labels | 400 labels | 2500 labels | 10000 labels |
| Π Model (2015) [58] | 75.56 ± 2.02 | 47.69 ± 1.52 | 12.80 ± 0.09 | 86.96 ± 0.80 | 58.80 ± 0.66 | 36.65 ± 0.00 |
| MeanTeacher (2017) [9] | 76.93 ± 2.29 | 56.06 ± 2.03 | 15.47 ± 0.43 | 81.11 ± 1.44 | 45.17 ± 1.06 | 31.75 ± 0.23 |
| FixMatch (2020) [15] | 8.33 ± 1.41 | 4.97 ± 0.05 | 4.29 ± 0.10 | 46.42 ± 0.82 | 28.03 ± 0.16 | 22.20 ± 0.12 |
| Dash (2021) [45] | 15.01 ± 3.70 | 5.13 ± 0.26 | 4.35 ± 0.09 | 44.82 ± 0.96 | 27.15 ± 0.22 | 21.88 ± 0.07 |
| MPL (2021) [44] | 6.62 ± 0.91 | 5.76 ± 0.24 | 4.55 ± 0.04 | 46.26 ± 1.84 | 27.71 ± 0.19 | 21.74 ± 0.09 |
| DoubleMatch (2022) [42] | 13.59 ± 5.60 | 5.56 ± 0.42 | 4.65 ± 0.17 | 41.83 ± 1.22 | 27.07 ± 0.26 | 21.22 ± 0.17 |
| NP-Match (2022) [60] | 4.91 ± 0.04 | 4.96 ± 0.06 | 4.11 ± 0.02 | 38.91 ± 0.99 | 26.03 ± 0.26 | 21.22 ± 0.13 |
| SoftMatch (2023) [47] | 5.06 ± 0.02 | 4.84 ± 0.10 | 4.27 ± 0.12 | 37.10 ± 0.77 | 26.66 ± 0.25 | 22.03 ± 0.03 |
| RefixMatch (2023) [59] | 4.94 ± 0.01 | 4.83 ± 0.05 | 4.18 ± 0.05 | 46.12 ± 1.07 | 27.28 ± 0.22 | 21.60 ± 0.04 |
| SimMatchV2 (2023) [21] | 4.90 ± 0.16 | 5.04 ± 0.09 | 4.33 ± 0.16 | 36.68 ± 0.86 | 26.66 ± 0.38 | 21.37 ± 0.20 |
| FullMatch (2023) [48] | 5.89 ± 1.01 | 4.64 ± 0.12 | 3.75 ± 0.08 | 40.58 ± 1.40 | 26.94 ± 0.40 | 21.44 ± 0.10 |
| OTAMatch (2024) [61] | 4.89 ± 0.09 | 4.85 ± 0.12 | 4.03 ± 0.08 | 37.96 ± 1.06 | 26.06 ± 0.48 | 21.02 ± 0.13 |
| CRMatch (2023) [16] | 10.13 ± 2.09 | 4.67 ± 0.02 | 3.81 ± 0.16 | 39.45 ± 1.69 | 25.43 ± 0.14 | 20.40 ± 0.08 |
| FreeMatch (2023) [17] | 4.97 ± 0.09 | 4.85 ± 0.10 | 4.14 ± 0.19 | 37.98 ± 0.42 | 26.47 ± 0.20 | 21.68 ± 0.03 |
| CRMatch-FSR (Ours) | 6.48 ± 0.95 | **4.49 ± 0.18** | **3.58 ± 0.06** | **36.14 ± 0.11** | **24.11 ± 0.15** | **19.78 ± 0.25** |
| FreeMatch-FSR (Ours) | **4.67 ± 0.03** | 4.57 ± 0.03 | 3.97 ± 0.07 | 36.61 ± 0.03 | 26.19 ± 0.46 | 21.45 ± 0.08 |

TABLE II

MEAN ERROR RATES (%) AND STANDARD DEVIATIONS ON SVHN AND STL-10 DATASETS WITH DIFFERENT LABEL BUDGETS. THE BEST RESULTS ARE HIGHLIGHTED IN BOLD AND THE SECOND-BEST RESULTS ARE UNDERLINED

| Dataset | SVHN | | | STL-10 |
|---|---|---|---|---|
| | 40 labels | 250 labels | 1000 labels | 1000 labels |
| Π Model (2015) [58] | 77.38 ± 5.36 | 13.55 ± 0.42 | 7.24 ± 0.19 | 32.78 ± 0.40 |
| MeanTeacher (2017) [9] | 81.94 ± 1.33 | 25.10 ± 3.17 | 12.29 ± 0.45 | 33.90 ± 1.37 |
| FixMatch (2020) [15] | 2.08 ± 0.08 | 1.99 ± 0.05 | 2.00 ± 0.04 | 6.25 ± 0.33 |
| Dash (2021) [45] | 2.08 ± 0.09 | 1.97 ± 0.01 | 2.03 ± 0.03 | 6.39 ± 0.56 |
| MPL (2021) [44] | 9.33 ± 8.02 | 2.29 ± 0.04 | 2.28 ± 0.02 | 6.66 ± 0.00 |
| DoubleMatch (2022) [42] | 15.37 ± 11.81 | 2.37 ± 0.35 | 2.10 ± 0.07 | **4.35 ± 0.20** |
| NP-Match (2022) [60] | 14.20 ± 0.67 | 9.51 ± 0.37 | 5.59 ± 0.24 | 5.59 ± 0.24 |
| SoftMatch (2023) [47] | 2.31 ± 0.00 | 2.15 ± 0.05 | 2.08 ± 0.04 | 5.73 ± 0.24 |
| RefixMatch (2023) [59] | 2.15 ± 1.23 | - | 1.89 ± 0.03 | 5.74 ± 0.30 |
| SimMatchV2 (2023) [21] | 7.92 ± 2.80 | 2.92 ± 0.81 | 2.85 ± 0.91 | 5.65 ± 0.26 |
| FullMatch (2023) [48] | 2.35 ± 0.10 | - | 1.99 ± 0.03 | 5.74 ± 0.09 |
| OTAMatch (2024) [61] | - | - | - | 5.70 ± 0.35 |
| CRMatch (2023) [16] | 7.14 ± 2.28 | 1.88 ± 0.04 | 1.91 ± 0.02 | 6.53 ± 0.36 |
| FreeMatch (2023) [17] | 3.79 ± 0.03 | 4.09 ± 0.66 | 4.31 ± 0.00 | 5.63 ± 0.15 |
| CRMatch-FSR | **1.86 ± 0.01** | **1.84 ± 0.02** | **1.80 ± 0.03** | **4.96 ± 0.03** |
| FreeMatch-FSR | 2.63 ± 0.09 | 3.65 ± 0.31 | 3.53 ± 0.06 | 5.48 ± 0.22 |

### *B. Results for Performance Comparison*

Performance comparisons between the FSR-augmented SSL methods (i.e. CRMatch-FSR and FreeMatch-FSR) and the other SSL baselines on the benchmarks were made, with the results shown in Tables I and II. It is obvious that the dual-branch FSR module, when inserted into CRMatch or FreeMatch, helped either model achieve significant performance improvements, especially in the cases where fewer labeled samples were used.

As illustrated in Table I, for CIFAR-10, CRMatch obtained a baseline error rate of 10.13% when only 4 labels per class (i.e. 40 labels in total) were available. With the FSR module added, the error rate of CRMatch reduced to 6.48% by 3.65%. In the experiments with 25 and 400 labels per class (i.e. 250 and 4000 labels in total), the error rate of CRMatch-FSR decreased by 0.18% and 0.23% respectively, compared with those of CRMatch. FreeMatch obtained error rates of 4.97%, 4.85%, and 4.14% in the three label budget scenarios respectively, yet the FSR module further reduced them to 4.67%, 4.57%, and 3.97%, respectively. Notably, among all the compared methods, CRMatch-FSR and FreeMatch-FSR achieved the best or near best results for this dataset.

Similar performance gains by the dual-branch FSR module can also be observed for CIFAR-100. For CRMatch, the error rates, with 4, 25, and 100 labels per class (i.e. 400, 2 500, and 10 000 in total) used, experienced noticeable declines from 39.45%, 25.43%, and 20.40% to 36.14%, 24.11%, and 19.78%, respectively, once the FSR module was introduced. FreeMatch-FSR, with the help of the FSR module, also had a noticeable performance improvement over the baseline FreeMatch, with reductions of 1.37%, 0.28%, and 0.23% in error rates under the three label budget constraints respectively. Overall, CRMatch-FSR outperformed almost all of its competitors, often achieving the top or runner-up position, while FreeMatch-FSR ranked close behind. This verifies the effectiveness of the dual-branch FSR module in feature learning for the more challenging dataset CIFAR-100.

The results in Table II show that the inclusion of FSR module again led to stable performance improvements of the base SSL models for SVHN and STL-10. As For CRMatch, when only 4 labels per class (40 labels total) were used, the error rate plunged from 7.14% to 1.86%, a remarkable reduction of 5.28%, thanking to the FSR module. In the experiments with 25 and 100 labels per class (250 and 1 000 labels in total), the error rates dropped from 1.88% to 1.84% and from 1.91% to 1.80%, respectively. FreeMatch also benefited from the FSR module. When 4 labels per class were available, the error rate decreased from 3.79% to 2.63% by 1.16%. In the cases of 25 and 100 labels per class, FreeMatch-FSR yielded error rates of 3.65% and 3.53% respectively, lower than those of the FreeMatch baseline (i.e. 4.09% and 4.31%) by 0.44% and 0.78% respectively. These results indicate that the dual-branch FSR module can still enhance the model's learning ability even when integrated into an SSL model with already good performance. For STL-10, with 100 labels per class (totally 1000 labels) available, the FSR module also conduced to the learning abilities of the base models. Specifically, CRMatch-FSR lowered the error rate from 6.53% (of CRMatch) to 4.96%, while FreeMatch-FSR achieved error rate reduction from 5.63% to 5.48%.

Compared with the recent existing semi-supervised approaches, CRMatch and FreeMatch have already demonstrated strong baseline performance on most of the datasets and different label budgets. Nevertheless, incorporation of the lightweight dual-branch FSR module not only yielded best-in-class or near-best performance in several cases (e.g., CIFAR-10 with 250/400 labels per class, and CIFAR-100 with 4 labels per class), but also brought considerable substantial performance gains in the few-label scenarios, such as CIFAR-10 and SVHN with only 4 labels per class. These findings make it confirmed that the FSR module provides a robust and broadly applicable enhancer to both contrastive-style (CRMatch) and confidence-based (FreeMatch) SSL frameworks, consistently improving accuracies of these already strong base models.

### *C. Ablation Study*

In order to further evaluate the proposed dual-branch FSR module, we performed ablation study on CIFAR-100 and STL-10. Since the FSR module contains two components, i.e. the dual-branch header and FSR block, the ablation study was undertaken for the following three variants based on CRMatch and FreeMatch.

**Baseline**: the original base SSL model, i.e. CRMatch or FreeMatch.

**Baseline+DBH**: the combination of the base SSL model and the dual-branch header without FSR block.

**Baseline+DBH with FSR**: the integration of the base SSL with the whole dual-branch FSR module.

Table III reports the Top-1 error rates of these three configurations on CIFAR-10 and SVHN. These results illustrate the effect of the dual-branch header (DBH) on the base SSL model and the additional performance gains from FSR.

As for the impact of the DBH on model performance, we can see from the results in Table III that by introducing the DBH alone into CRMatch or FreeMatch, the model performance did not improve in all the cases; in some cases, it even deteriorated. In the case of CIFAR-10 with only 4 labels per class available, with the DBH, the error rate of CRMatch increased from 10.13% to 14.94%. When 400 labels per class were used, a minor performance enhancement can be observed, with the error rate decreasing from 3.81% to 3.68%.

On SVHN, however, the DBH alone can make the error rate of CRMatch have a sharp decrease from 7.14% to 1.89% with 4 labels per class available. These results suggest that the effect of DBH is highly case-dependent. For FreeMatch, insertion of the DBH alone brought minor performance improvement in most of the cases except for SVHN with 25 labels per class available. The above comparison confirms that the performance improvements brought about by the DBH without FSR are limited and highly dependent on datasets and label budget constraints.

TABLE III
ABLATION RESULTS OF THE DBH AND FSR COMPONENTS ON CIFAR-10 AND SVHN. THE BEST RESULTS ARE HIGHLIGHTED IN BOLD AND THE SECOND-BEST RESULTS ARE UNDERLINED

| Methods | CIFAR-10 | | | SVHN | | |
|---|---|---|---|---|---|---|
| | 40 labels | 250 labels | 4000 labels | 40 labels | 250 labels | 1000 labels |
| **CRMatch** | 10.13 ± 2.09 | 4.67 ± 0.02 | 3.81 ± 0.16 | 7.14 ± 2.28 | 1.88 ± 0.04 | 1.91 ± 0.02 |
| **CRMatch + DBH** | 14.94 ± 3.86 | 4.86 ± 0.27 | 3.68 ± 0.09 | 1.89 ± 0.05 | 1.86 ± 0.05 | 1.85 ± 0.03 |
| **CRMatch + DBH with FSR** | **6.48 ± 0.95** | **4.49 ± 0.18** | **3.58 ± 0.06** | **1.86 ± 0.01** | **1.84 ± 0.02** | **1.80 ± 0.03** |
| **FreeMatch** | 4.97 ± 0.09 | 4.85 ± 0.10 | 4.14 ± 0.19 | 3.79 ± 0.03 | 4.09 ± 0.66 | 4.31 ± 0.00 |
| **FreeMatch + DBH** | 4.71 ± 0.03 | 4.69 ± 0.08 | 4.08 ± 0.15 | 3.51 ± 0.29 | 4.21 ± 0.66 | 4.01 ± 0.28 |
| **FreeMatch + DBH with FSR** | **4.67 ± 0.03** | **4.57 ± 0.03** | **3.97 ± 0.07** | **2.63 ± 0.09** | **3.65 ± 0.31** | **3.53 ± 0.06** |

When the whole dual-branch FSR module was inserted into the base SSL framework, with the assistance from FSR in model training, consistent performance improvements across all the settings can be observed. CRMatch+DBH with FSR, essentially CRMatch-FSR in Tables I and II, obtained an error rate of 6.48%, substantially lower than that of both CRMatch (10.13%) and CRMatch+DBH (14.94%). When 25 labels were used, FSR promoted performance improvement of CRMatch+DBH, with its error rate decreasing remarkably to lower than CRMatch by 0.18%. In the other cases, including those for SVHN, although CRMatch+DBH had better accuracies than the base model, FSR helped to further improve the classification accuracies. When the base SSL model was FreeMatch, the DBH-only variant yielded only marginal performance gains and even performance degradation in some cases (e.g. SVHN with 25 labels per class). However, with the help of FSR in model training, FreeMatch+DBH with FSR had further error rate reductions in all the cases. Especially, for SVHN with 4 labels per class, FreeMatch+DBH with FSR had the error rate decreased from 3.51% of FreeMatch+DBH to 2.63%, which, compared with 3.79% of FreeMatch, had a 1.16% drop.

The above ablation study reveals the following facts. First, the influence of DBH alone on the model performance is sensitive to the datasets and the label budgets, and even can be detrimental in the cases with very few labels. Second, the FSR mechanism, when introduced into the dual-branch module for model training, can consistently enhance the representation ability of the baseline+DBH, resulting in robust model performance gains, especially in the cases with few labels.

Fig. 3 visualizes the evolution of top-1 accuracy over $2^{20}$ training iterations in the cases where 400 labels per class were available for CIFAR-10 and SVHN. Fig.3(a) shows the results obtained with FreeMatch being the base model for CIFAR-10, whereas Fig. 3(b) presents the evolution process with CRMatch being the base model for SVHN. In either case, the top-1 accuracy curve of baseline+DBH with FSR (the red line) stays above that of baseline+DBH (the blue line) during the training process, which means that baseline+DBH with FSR has higher accuracies at nearly all the checkpoints than baseline+DBH.

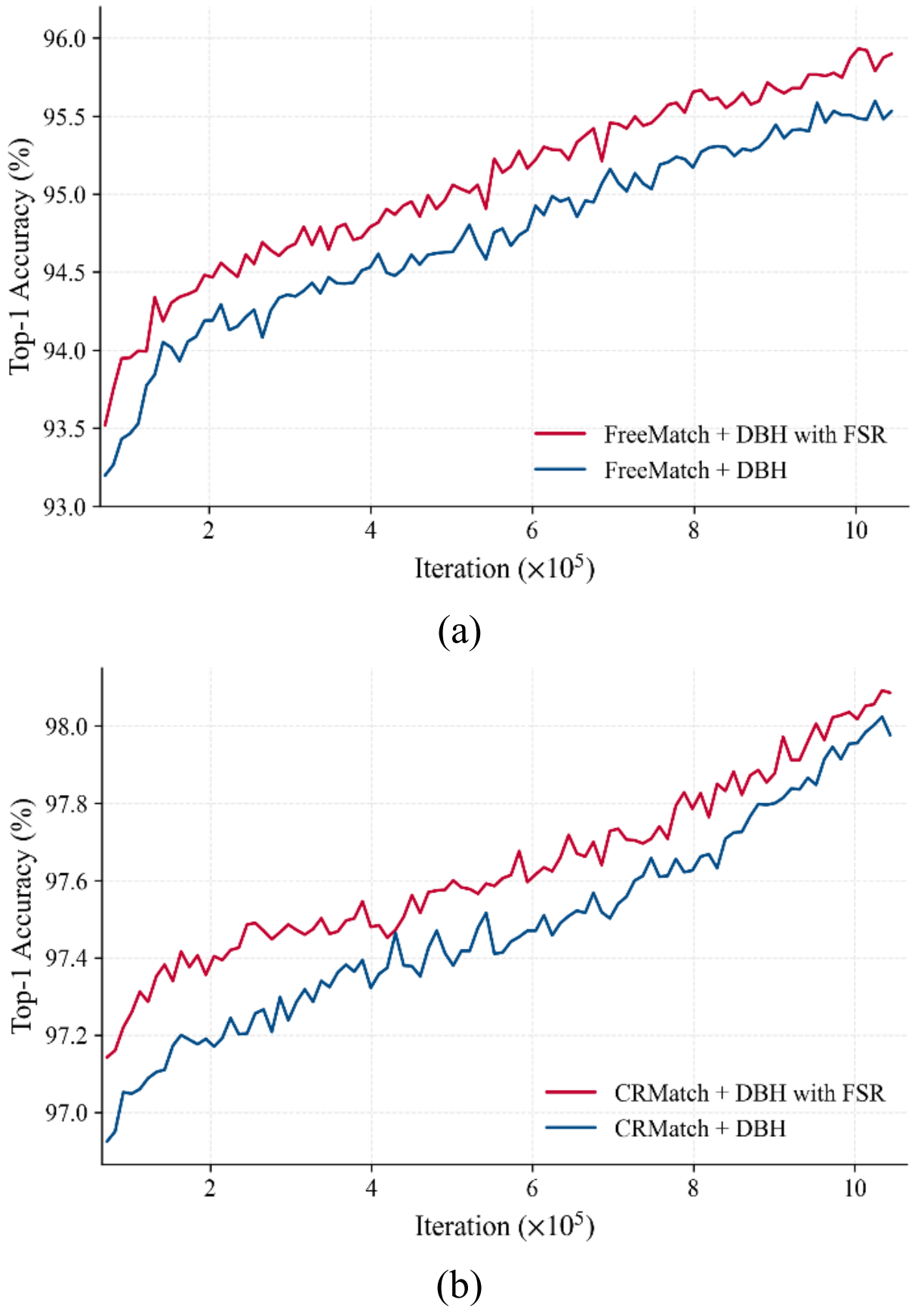

**Fig. 3.** Comparison evolution of Top-1 accuracies between DBH only and DBH with FSR. (a) CIFAR-10 with 400 labels per class in cases of FreeMatch. (b) SVHN with 100 labels per class in cases of CRMatch base. The curves illustrate the consistent performance gains provided by the FSR module during training.

We further examined the influence of FSR on the learned feature representation by projecting the strongly augmented unlabeled feature samples to a two-dimensional space with t-SNE. Fig. 4 shows the projections of feature representations learned by the two variants of FreeMatch for CIFAR-10 with 400 labels per class available (i.e. 4 000 labeled images in total). It is shown by Fig. 4(a) that in the feature embeddings of FreeMatch+DBH, the feature samples of several classes overlap or are spread unevenly over the plane. However, as shown in Fig. 4(b), with FSR employed for training FreeMatch+DBH, the feature sample points form more compact intra-class groups and exhibit clearer inter-class separation, and the clusters are distributed more uniformly over the embedding space. This implies that the feature representation learned with the help of FSR has better discriminability. These qualitative patterns align with the quantitative results reported previously, and illustrate how FSR reshapes the distribution of features extracted by the model for the unlabeled images.

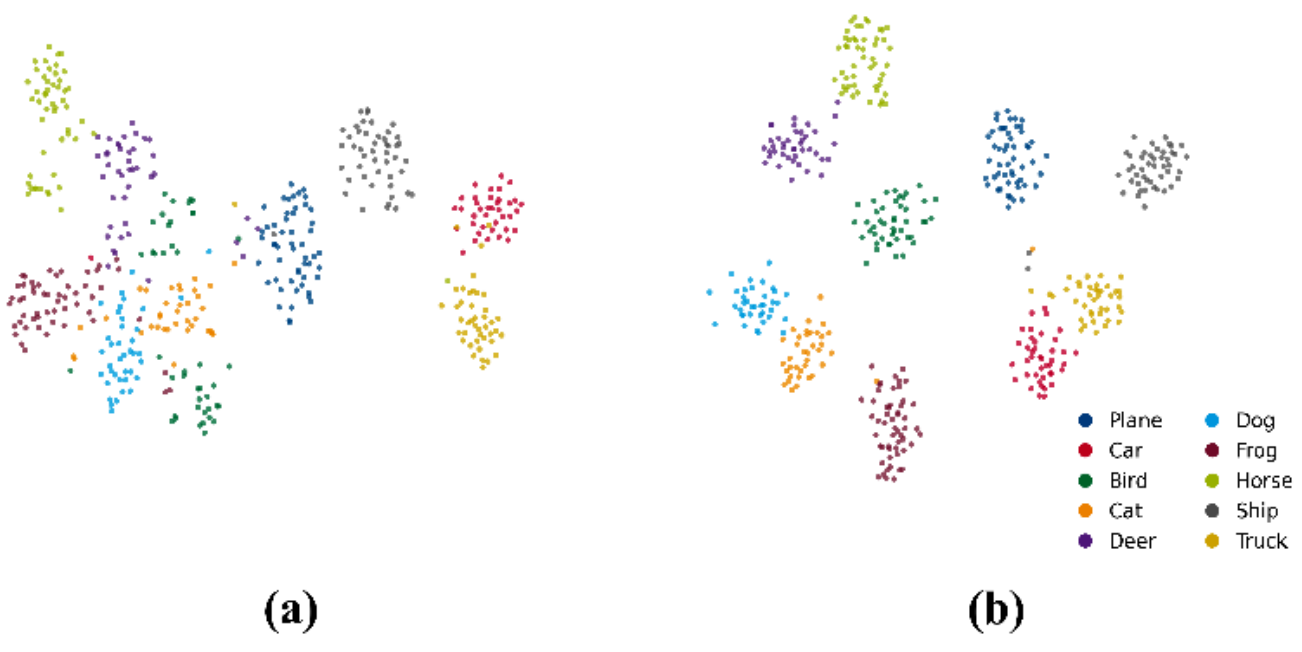

**Fig. 4.** t-SNE visualization of feature embeddings for strongly augmented unlabeled samples on CIFAR-10 with 400 labels per class available and FreeMatch being base SSL. Fig.4(a) shows that of the baseline + DBH variant, whereas Fig.4(b) visualizes that of baseline+DBH with FSR. It is shown that with the additional FSR block, the obtained embeddings by the model form tighter intra-class clusters and exhibit larger inter-class separation, and the clusters distribute over the embedding space more uniformly.

*D. Selection of Hyperparameters*

There are three hyperparameters in the loss function of the FSR-augmented SSL framework, i.e. $\lambda_u$, $\lambda_b$, and $\lambda_r$. $\lambda_u$ is the weight of pseudo-labeling loss and was set to $\lambda_u = 1$ in all of our experiments as in the existing base SSL framework. However, $\lambda_b$ and $\lambda_r$ are the additional hyperparameters brought by our proposed FSR module. Thus, in this work, we only investigated how to select these two hyperparameters to make the learned model have generally good performance.

We selected several representative value pairs for $\lambda_b$ and $\lambda_r$. Experiments were carried out on CIFAR-10 (25 labels per class) by using FreeMatch+DBH with FSR, i.e. FreeMatch-FSR, and the resulting error rates for different value pairs of $\lambda_b$ and $\lambda_r$ are summarized in Table IV. When both $\lambda_b$ and $\lambda_r$ were zero, the error rate was 4.75%. In this case, the last two terms of the FSR loss in equation (21) vanished, and the FSR constraints did not actually work in the absence of these two terms. When both $\lambda_b$ and $\lambda_r$ became nonzero, FSR took effect in model training with the performance of the trained model improved. However, larger values of these two hyperparameters are not necessarily better. For example, if $\lambda_b = \lambda_r = 0.1$, the resulting error rate was 4.65%, although better than that of FreeMatch+DBH without FSR. The lowest error rate was achieved when $\lambda_b = 0.01$ and $\lambda_r = 0.001$. Over all the tested value pairs of $\lambda_b$ and $\lambda_r$, the error rate varied within the range of only about 0.2%, which suggests that the model performance is not very sensitive to $\lambda_b$ and $\lambda_r$ as long as they are not zero. We therefore set $\lambda_b = 0.01$ and $\lambda_r = 0.001$ as defaults in all of the other experiments in this work.

TABLE IV

SENSITIVITY ANALYSIS OF $\lambda_b$ AND $\lambda_r$ ON CIFAR-10 (25 LABELS PER CLASS AVAILABLE) WITH FREEMATCH BEING THE BASE SSL FRAMEWORK

| $\lambda_b$ | $\lambda_r$ | Error Rate (%) ↓ |
|---|---|---|
| 0.0 | 0.0 | 4.75 |
| 0.0 | 0.001 | 4.69 |
| 0.001 | 0.001 | 4.63 |
| 0.01 | 0.0 | 4.72 |
| 0.01 | 0.01 | 4.63 |
| 0.01 | 0.1 | 4.64 |
| 0.1 | 0.001 | 4.59 |
| 0.1 | 0.1 | 4.65 |
| **0.01** | **0.001** | **4.54** |

*E. Complexity Analysis*

The computational complexity of the proposed framework was analyzed in terms of run-time cost per iteration, GPU memory used per iteration, and the number of additional parameters introduced by the dual-branch FSR module. The results are summarized in Table V and Table VI. Table V reports the peak GPU memory (MB) and wall clock time per training iteration (ms) for the variants of CRMatch and FreeMatch on the four datasets. As can be seen, the introduction of the dual-branch header (DBH) increased the model's GPU memory requirements by at most 2 MB and incurred an additional latency of 0.1 ms, which is well within the run-to-run noise of the RTX 4090. By adding the FSR block, DBH + FSR had an equally negligible impact on the model in computational complexity. For instance, on CIFAR-10, the time consumption per iteration for training CRMatch marginally shifted from 12.58 ms to 12.51 ms, and the GPU memory usage remained practically unchanged. The observation is consistent across all the other datasets including CIFAR-100, SVHN, and STL-10. This indicates that the proposed modules did not incur any perceivable overhead in terms of training time or GPU memory.

TABLE V

GPU MEMORY USAGE AND TRAINING TIME PER ITERATION OF EACH VARIANT OF CRMATCH OR FREEMATCH ON FOUR DATASETS (BACKBONE MODELS IN PARENTHESES)

| Methods | CIFAR10 (WRN28-2) | | CIFAR100 (WRN28-8) | | SVHN (WRN28-2) | | STL-10 (WRN37-2) | |
|---|---|---|---|---|---|---|---|---|
| | Memory | Time | Memory | Time | Memory | Time | Memory | Time |
| **CRMatch** | 2922 MB | 12.58 ms | 10268 MB | 25.79 ms | 2923 MB | 16.48 ms | 8102 MB | 20.61 ms |
| **CRMatch+DBH** | 2924 MB | 12.41 ms | 10270 MB | 25.81 ms | 2924 MB | 16.44 ms | 8112 MB | 20.64 ms |
| **CRMatch+DBH with FSR** | 2924 MB | 12.51 ms | 10274 MB | 25.85 ms | 2926 MB | 16.47 ms | 8110 MB | 20.62 ms |
| **FreeMatch** | 2500 MB | 10.69 ms | 9939 MB | 20.95 ms | 2501 MB | 14.34 ms | 7095 MB | 17.45 ms |
| **FreeMatch+DBH** | 2501 MB | 10.79 ms | 9942 MB | 21.03 ms | 2502 MB | 14.34 ms | 7100 MB | 17.41 ms |
| **FreeMatch+DBH with FSR** | 2502 MB | 10.76 ms | 9944 MB | 21.10 ms | 2502 MB | 14.38 ms | 7104 MB | 17.43 ms |

Table VI lists the dimensionality of each branch in the DBH for each backbone model and the additional number of parameters introduced by the DBH and the FSR block with CRMatch and FreeMatch being base SSL frameworks respectively. Here, $d = \frac{1}{2}D$ is the dimensionality of feature vectors output by each branch of the DBH, where $D$ is that of feature vector output by the backbone. It is shown that in each case of CRMatch, the ratio of the number of parameters introduced by the dual-branch FSR module (DBH + FSR) to the total number of parameters is below 1%; and in each case of FreeMatch, the ratio is below 1.5%. Hence, the enhancement of the model's representation ability was achieved by the FSR module with an imperceptible increase of parameters.

TABLE VI

THE PARAMETER NUMBERS OF THE DBH AND THE FSR FOR DIFFERENT DATASETS AND BACKBONES IN CRMATCH AND FREEMATCH CASES

CRMatch

| Dataset | Backbone | $d$ | DBH Params | FSR Params | Total Params | Ratio |
|---|---|---|---|---|---|---|
| CIFAR-10 | WRN-28-2 | 64 | 16,640 | 4,160 | 2,554,158 | 0.81% |
| CIFAR-100 | WRN-28-8 | 256 | 263,168 | 65,792 | 40,772,424 | 0.81% |
| SVHN | WRN-28-2 | 64 | 16,640 | 4,160 | 2,554,158 | 0.81% |
| STL-10 | WRN-37-2 | 128 | 66,048 | 16,512 | 8,438,382 | 0.98% |

FreeMatch

| Dataset | Backbone | $d$ | DBH Params | FSR Params | Total Params | Ratio |
|---|---|---|---|---|---|---|
| CIFAR-10 | WRN-28-2 | 64 | 16640 | 4 160 | 1 488 426 | 1.40% |
| CIFAR-100 | WRN-28-8 | 256 | 263168 | 65 792 | 23 729 988 | 1.39% |
| SVHN | WRN-28-2 | 64 | 16640 | 4 160 | 1 488 426 | 1.40% |
| STL-10 | WRN-37-2 | 128 | 66048 | 16 512 | 6 012 010 | 1.37% |

## V. CONCLUSION

Although SSL has made rapid progress recently, unfortunately, most of the advancement has been made at the cost of increasingly complicated learning algorithms with sophisticated data perturbation methods used in consistency regularization or pseudo-labeling. In this paper, we proposed the feature space renormalization (FSR) mechanism, and then a dual-branch FSR module for SSL. The FSR mechanism, based on the group representation theory, exerts consistency at the feature level to enable the model to learn good feature representation. The dual-branch FSR module, the combination of a dual-branch header and an FSR block, can be integrated as plug-and-play module into the existing SSL frameworks to improve the discrimination ability of the base SSL models. The proposed FSR module was evaluated on four widely used benchmark datasets by inserting it into two state-of-the-art SSL approaches, i.e. CRMatch and FreeMatch. It was shown that, both CRMatch and FreeMatch, with the help of the FSR module, achieved significant performance gains with negligible additional overhead in terms of computation time and GPU memory.

Although only CRMatch and FreeMatch were used as the base SSL frameworks in this work, the FSR module is a generic and plug-and-play module that can be seamlessly inserted into other SSL frameworks without altering their data pipelines or loss functions. Our future work will explore coupling FSR with larger backbones, vision transformers and multi-modal pre-trained models, aiming to narrow the remaining gap between semi-supervised and fully supervised learning under realistic resource budgets.

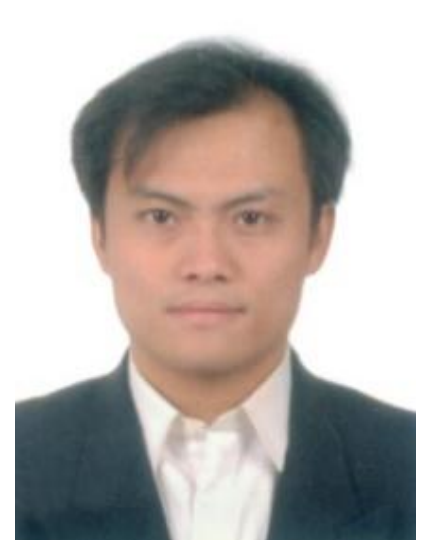

**Jun Sun** (Member, IEEE) received his PhD in control theory and engineering, and an MSc in Computer Science and Technology from Jiangnan University, China, in 2009 and 2003, respectively. He is currently working as a full Professor with the Department of Computer Science and Technology, Jiangnan University, China. He is also vice director of Jiangsu Provincial Engineering Laboratory of Pattern Recognition and Computational Intelligence, Jiangsu Province. His major research areas and work are related to computational intelligence, machine learning, bioinformatics, among others. He published more than 150 papers in journals, conference proceedings and several books in the above areas.

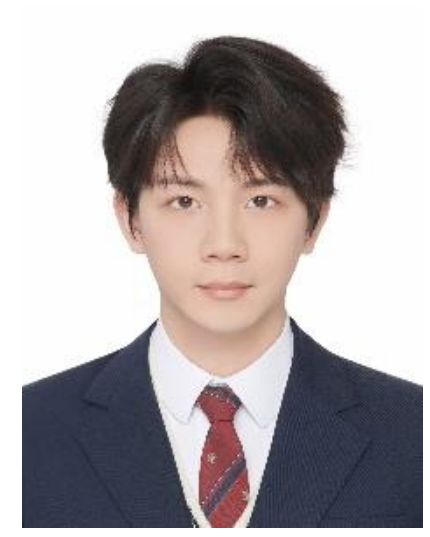

**Wancheng Zhang** is currently pursuing the M.Eng. degree in Computer Science and Technology with the School of Artificial Intelligence and Computer Science, Jiangnan University, Wuxi, China. His research interests include co-mputer vision and machine learning, with a particular focus on semi-supervised learning and representation learning.

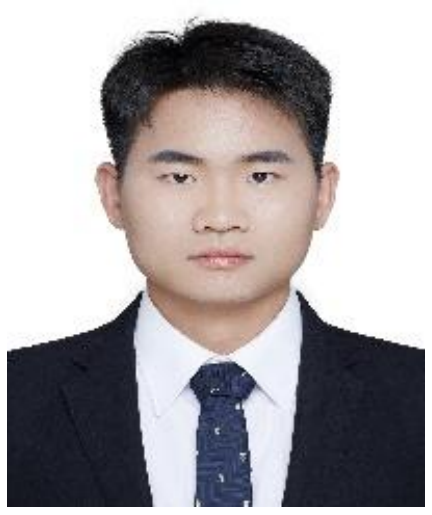

**Chao Zhou** studied for a master's degree in computer science and technology at the School of Artificial Intelligence and Computer Science in Jiangnan University, China, in 2020. He is currently working toward the PhD degree in software engineering in the Jiangsu Provincial Engineering Laboratory of Pattern Recognition and Computational Intelligence, Jiangnan University, Wuxi, China. His research interests include computer vision, representation learning and incremental learning.

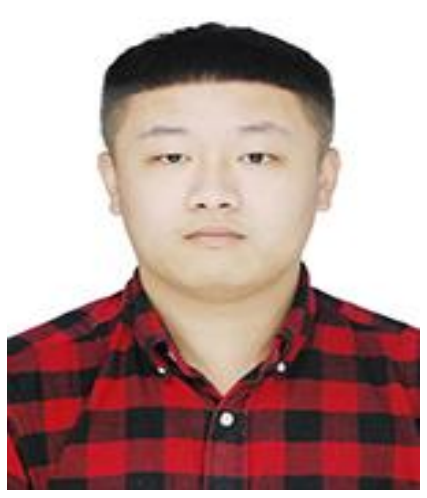

**Zhongjie Mao** studied for a master's degree in computer science and technology at the School of Internet of Things Engineering in Jiangnan University, China, in 2016. He is currently working toward the PhD degree in pattern recognition and intelligent systems in the Jiangsu Provincial Engineering Laboratory of Pattern Recognition and Computational Intelligence, Jiangnan University, Wuxi, China. His research interests include computer vision, semi-supervised learning and representation learning.

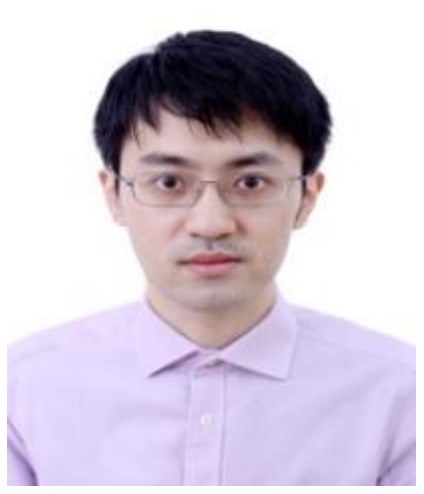

**Chao Li** received his Bachelor Degree in Information Engineering from Nanjing University of Information Science and Technology, Nanjing, China, in 2012 and a Master Degree in Computer Science and Technology from Jiangnan University, Wuxi, China, in 2017. He is currently working as a Ph.D. student in Control Science and Engineering in Jiangnan University, Wuxi, China. His research areas are related to computational intelligence, computer vision and bioinformatics. He has published several papers in journals and conferences in the above areas.

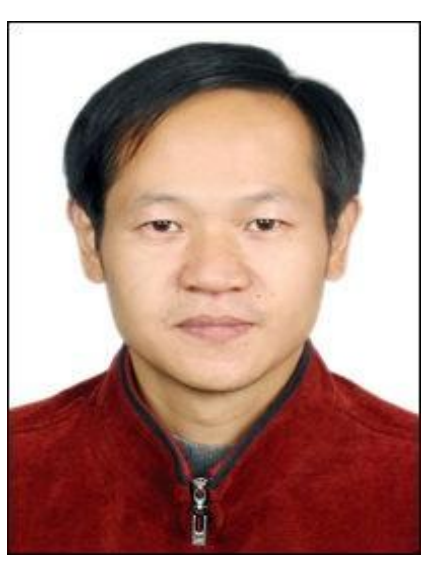

**Xiao-Jun Wu** received his B.Sc. degree in mathematics from Nanjing Normal University, Nanjing, China, in 1991. He received the M.S. degree in 1996, and the Ph.D. degree in pattern recognition and intelligent systems in 2002, both from Nanjing University of Science and Technology, Nanjing, China. He joined Jiangnan University in 2006, where he is currently a professor. He has published more than 200 papers in his fields of research. He was a visiting researcher in the Centre for Vision, Speech, and Signal Processing (CVSSP), University of Surrey, U.K., from 2003 to 2004. His current research interests include pattern recognition, computer vision, fuzzy systems, neural networks, and intelligent systems.